\documentclass{article}
\usepackage{spconf, amsmath, graphicx, amssymb}
\usepackage{algorithmicx}
\usepackage[ruled]{algorithm}
\usepackage[noend]{algpseudocode}
\usepackage{subfig}

\def\G{\mathcal{G}} 
\def\E{\mathcal{E}} 
\def\L{\mathcal{L}} 
\def\V{\mathcal{V}} 

\DeclareMathOperator*{\argmin}{arg\,min}

\copyrightnotice{978-1-5090-0746-2/16/\$31.00 {\copyright}2016 IEEE}

\title{SOURCE LOCALIZATION ON GRAPHS VIA $\ell_1$ RECOVERY AND SPECTRAL GRAPH THEORY}
\name{Rodrigo Pena, Xavier Bresson, Pierre Vandergheynst}
\address{Ecole Polytechnique Federale de Lausanne (EPFL)\\
School of Engineering (STI)\\
}
\begin{document}
%
\maketitle
\begin{abstract}
We cast the problem of source localization on graphs as the simultaneous problem of sparse recovery and diffusion kernel learning. An  $\ell_1$ regularization term enforces the sparsity constraint while we recover the sources of diffusion from a single snapshot of the diffusion process. The diffusion kernel is estimated by assuming the process to be as generic as the standard heat diffusion. We show with synthetic data that we can concomitantly learn the diffusion kernel and the sources, given an estimated initialization. We validate our model with cholera mortality and atmospheric tracer diffusion data, showing also that the accuracy of the solution depends on the construction of the graph from the data points.
\end{abstract}
\begin{keywords}
Source localization, graph, sparsity, optimization
\end{keywords}
%
\vspace{-1.5em}
\section{Introduction}
\label{sec:intro}
\vspace{-0.5em}
Source localization vaguely refers to a wide class of problems in which the spatial origins of some given diffused information are important to identify. Finding the starting point of an epidemic, the source of heat in a sensor network, or the origin of a rumor in a social network are all examples that fit into this category. We aim at introducing an abstract framework for solving this class of problems without knowledge of the number of sources, and with soft assumptions on the information diffusion process. Our framework leverages the structure of the signal to be recovered, namely its sparsity.

An important source of inspiration for our work is the research of Cand\`es and Fernandez-Granda on the super-resolution of point sources \cite{FernandezGranda:2015uc}, \cite{Candes:2012uf}, \cite{Candes:2013wu}. They study the estimation of sparse signals, with support in a subset of $\mathbb{R}$, from low-resolution observations. The measurement, $y$, is modeled as a convolution of the original sparse signal, $x$, with a low-pass point-spread function, and the recovery of $x$ from $y$ is cast as a convex optimization problem, with a fidelity term and a sparsity-inducing norm on $x$. Cand\`es and Fernandez-Granda show that, in the noiseless setting, $x$ can be exactly recovered from $y$ by solving this optimization problem, as long as the spikes in $x$ obey a certain minimum separation constraint \cite{FernandezGranda:2015uc}. This technique has also been studied in detail by Duval and Peyr\'e in \cite{Duval:2015}.

Our attempt in this work is to cast a similar optimization problem to solve source localization problems on \textit{graphs}. We do this by modeling the source signals as functions whose domain consists on the nodes of the network. In this context, the eigenvectors  of the graph Laplacian play a similar role as the Fourier modes on the real line, and the diffusion of the source signals can be modeled as the action of a linear operator which is a function of this graph Laplacian \cite{Shuman:2013et}. Unlike Cand\`es and Fernandez-Granda, however, we do not assume in general that the diffusion process is known. Rather, we assume it to be given by a parametrized function of the graph Laplacian and attempt to learn this parameter at the same time as the source locations. We note, however, that this simultaneous learning makes the overall optimization problem non-convex and initialization-dependent.

Unlike our proposed technique, which relies on a \textit{global} approach of diffusion, most other works in the literature focus on \textit{local} strategies, observing small fractions of nodes, and leveraging information from the detection times at the observed nodes. One such example is the work of Pinto \textit{et al.} \cite{Pinto:2012tp}, who propose a maximum likelihood estimator that is optimal for trees, and otherwise performs best on scale-free networks. Similarly, Feizi \textit{et al.} \cite{Feizi:2014tb} use maximum likelihood and minimum error estimators to identify the sources, but they improve on the complexity of the algorithm by modeling the diffusion of information among pairs of nodes as depending only on $k$-shortest-paths between them, which can be too strong an assumption in some cases. More recently, Zhang \textit{et al.} \cite{Zhang:2016wr} proposed a nonconvex regression learning model for estimating anomalous diffusion sources. They jointly learn the number of sources, and the propagation time and paths by observing the values and detection times on a subset of network nodes. We differ from Zhang \textit{et al.} by casting a different optimization problem, and by observing only a single snapshot of the diffusion process. We should also finally mention NETSLEUTH, by Prakash \textit{et al.} \cite{Prakash:2012uk}, that differs from the aforementioned strategies by employing the Minimum Description Length (MDL) principle to identify the set of source nodes.

Our main contributions can be summarized as follows:
\begin{itemize}
\vspace{-1em}
\item A generic optimization framework, made possible by using spectral graph theory tools;
\vspace{-1em}
\item Simultaneous learning of sparse sources and diffusion kernel from a single snapshot of the process;
\vspace{-0.75em}
\item Specification of an error measure for comparing the recovered sources to the ground truth; and
\vspace{-0.75em}
\item A first analysis of how the graph construction can influence the accuracy of the source localization.
\end{itemize}  


\vspace{-1.5em}
\section{Theory}
\label{sec:prob_setup}
\vspace{-0.5em}
We consider undirected, weighted graphs $\G = \left(\V, \E, W\right)$, consisting of a set of nodes $\V$, a set of edges $\E$, and a weighted adjacency matrix $W$. Each entry $W_{ij}$ of $W$ represents the weight of the edge between nodes $i,j \in \V$, with $W_{ij} = 0$ if vertices $i$ and $j$ are not connected. Because $\G$ is undirected, $W$ is a symmetric matrix. The sparse signal representing the sources of diffusion is a function $x: \V \to \mathbb{R}$ with support in a subset of $\V$.

Let $D$ be a diagonal matrix with entries $D_{ii} = \sum_{j}W_{ij}$, and call it the graph's degree matrix. The normalized graph Laplacian \cite{Chung:2001ud} is defined then as $\L = I - D^{-1/2}WD^{-1/2}$. By construction, the graph Laplacian is symmetric and positive semidefinite. Therefore, it admits an eigendecomposition, with non-negative eigenvalues, $\L = U \Lambda U^{T}$, where each column of $U$ is one of $\L$'s eigenvectors, \textit{i.e.}, the graph Fourier modes, and $\Lambda$ is a diagonal matrix whose entries are the eigenvalues corresponding to each of the eigenvectors in $U$. We assume, without loss of generality, that the eigenvalues in $\Lambda$ are ordered, \textit{i.e.}, $0 = \lambda_{0} \leq \lambda_{1} \leq ... \leq \lambda_{n} = \lambda_{max} \leq 2$, inducing a respective ordering on the columns of $U$.

We can model the diffusion of the sparse signal $x$ on the graph as the left-multiplication of a function of the Laplacian. The diffusion operator can also be defined on the spectral domain, as a function of the Laplacian eigenvalues. Throughout this work, we will use the heat kernel
\begin{equation}
\begin{matrix}
g_{\theta}(\lambda) = \exp\left(-\theta\lambda\right), &  \theta > 0.
\end{matrix}
\label{eq:heat_kernel}
\end{equation}
to model information diffusion, but the extension to other parametric diffusion kernels is straightforward. We obtain the corresponding diffusion matrix by simply returning from the spectral domain to the graph domain: $A_{\theta} = g_{\theta}(L) = Ug_{\theta}(\Lambda)U^{T}$.

We can finally state our optimization problem for source localization on graphs as
\begin{equation}
\underset{x, \theta}{\min} \, E(x, \theta) =  \underset{x, \theta}{\min} \, \left \{ \gamma \left |\left | x \right |\right |_{1} + \dfrac{\alpha}{2}\left |\left | A_{\theta}x - b \right |\right |_{2}^{2} \right \},
\label{eq:opt_problem}
\end{equation}
where $b$ is the observed signal on the graph. We promote sparsity on the sources through the $\ell_1$ norm, while the other term accounts for the fidelity with respect to the observations. The parameters $\gamma$ and $\alpha$ control the trade-off between those two.

In problem \eqref{eq:opt_problem} we have both $x$ and $\theta$ as unknowns, so we take an alternating approach to solving it. At iteration $k$, we optimize first for $x$ and then for $\theta$:
\begin{equation}
\left\{\begin{matrix}
x_{k + 1} = \underset{x}{\argmin} \, E(x, \theta_{k})\\ 
\theta_{k + 1} = \underset{\theta}{\argmin} \, E(x_{k+1}, \theta)
\end{matrix}\right.,
\label{eq:alt_opt_problem}
\end{equation}
given some initial point $(x_{0}, \theta_{0})$. We stop the process when $\left | E(x_{k+1}, \theta_{k+1}) - E(x_{k}, \theta_{k}) \right | < \epsilon$, for some fixed tolerance $\epsilon > 0$, or if it has attained a given maximum number of iterations. 

The first step of \eqref{eq:alt_opt_problem} is solved by fast iterative shrinkage-thresholding (FISTA) \cite{Beck:2009:FIS:1658360.1658364}, while the second step is solved with a smoothed version of Newton's method (simply adding a proximal term w.r.t. previous estimates $\theta_k$). The algorithms were implemented in MATLAB and are available in the first author's github repository 
\footnote{https://github.com/rodrigo-pena/src-localization-graphs} 

\vspace{-1.0em}
\subsection{Error measure}
\label{sec:err_meas}
\vspace{-0.5em}
We specify an error measure based on hop distances similar to the one given in \cite{Zhang:2016wr}. Let $x : \V \to \mathbb{R}$ be the reference signal, with non-zero values (spikes) on the source nodes, and let $y : \V \to \mathbb{R}$ be the test signal. Let also $\mathcal{A} \subseteq \V$, which we call the active nodes set, contain the nodes with spikes in $x$. For each $i \in \mathcal{A}$, define a set $\mathcal{N}_{i} \subseteq \V$ containing the nodes in $\V$ which are closer to $i$ than to any other element of $\mathcal{A}$. Call the set $\mathcal{N}_{i}$ the influence zone of node $i$. The distance $h(i, j)$ between two nodes $i, j \in \V$ is measured, in hops, as the shortest path in $\G$ between $i$ and $j$. The average hop error between those signals can then be written as
\begin{equation}
e(x, y) = \underset{i \in \mathcal{A}}{\sum}\frac{\underset{j \in \mathcal{N}_{i}}
{\sum}\left| y(j) \right| h(i, j)}
{\underset{j \in \mathcal{N}_{i}}
{\sum}\left| y(j) \right|}.
\label{eq:error_measure}
\end{equation}

Each term inside the outermost sum in \eqref{eq:error_measure} can be seen as the center of mass of $y$ in the influence zone of an active node $i$, when the origin of the coordinate system is set to node $i$. This interpretation makes clear that \eqref{eq:error_measure} penalizes both non-sparse test signals, and sparse, but misplaced (with respect to the ground-truth sources) spike signals. As a special case, when $y \equiv 0$ but $x \neq 0$, we set $e(x, y) = \infty$.  

\vspace{-1.0em}
\section{Experiments}
\label{sec:results}
\vspace{-1.0em}
\subsection{Sensor Graph}
\label{sec:sensor_graph}
\vspace{-0.5em}
Sensor graphs are constructed by first picking random points on the plane and then connecting each one to its k-nearest neighbors ($k$-NN). The weight of the edge between two nodes, $i$ and $j$, is given by $exp(\frac{-d(i, j)^{2}}{\sigma^{2}})$, where $d(i,j)$ is the Euclidean distance between their respective coordinates, and $\sigma^{2}$ is a scaling factor.

\begin{figure}[htb]
	\vspace{-1em}
	\centering
	\subfloat[Average hop error as a function of source distance and diffusion time $\theta$.]{\includegraphics[width=0.49\linewidth]{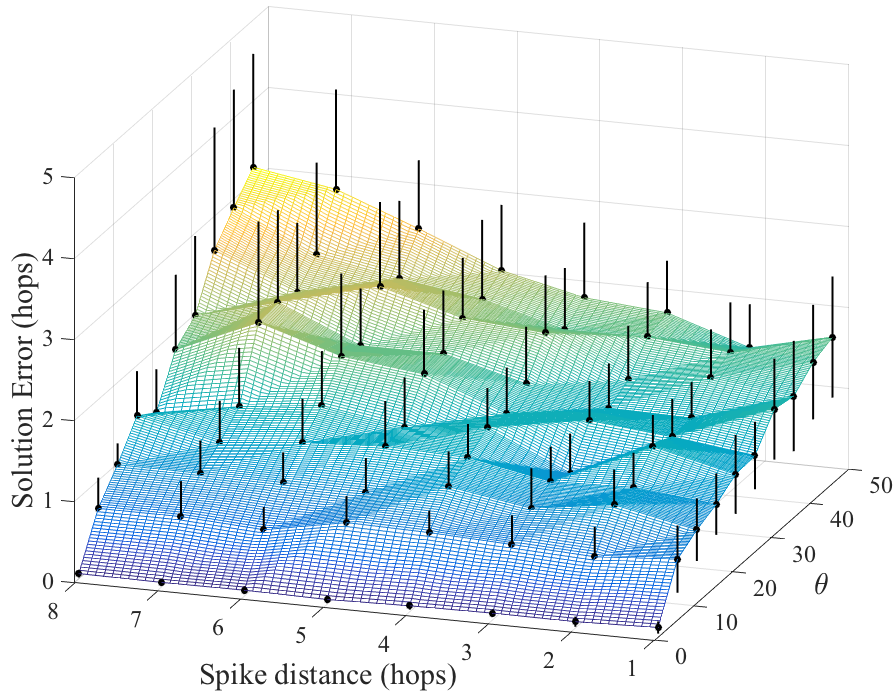}\label{fig:spike_distance}}
	~
	\subfloat[Average hop error as a function of the SNR and the diffusion time $\theta$.]{\includegraphics[width=0.49\linewidth]{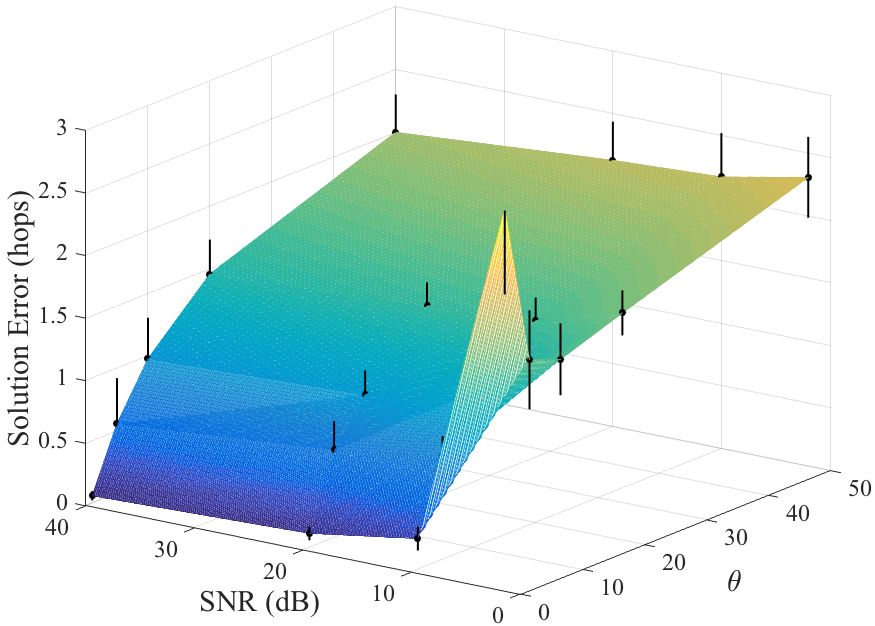}\label{fig:noise}}	
	\vspace{-0.5em}
\caption{Experiments on the sensor graph.}
\end{figure}

We first  analyze the accuracy of the solution to \eqref{eq:opt_problem} with respect to both spike distances, $h$, and diffusion times, $\theta$. For each pair $(h, \theta)$, and for each trial, we first pick at random two spikes, $h$ hops away from one another, on a $250$-node sensor graph. We then diffuse this spike signal using \eqref{eq:heat_kernel} with parameter $\theta$ to obtain observation $b$. We recover a sparse signal from this observation by solving \eqref{eq:opt_problem}, and measure the error  of this solution with respect to the originally drawn spikes. Figure \ref{fig:spike_distance} shows the average and standard deviations, over 32 trials, of the hop error \eqref{eq:error_measure} of the recovered solution for different pairs $(h, \theta)$. We see that the hop distance between the sources of diffusion does not seem to affect the accuracy of the solution, while the diffusion time $\theta$ has a lot of influence on it.
  
In our next experiment, we analyze the influence of noise. At each trial, we randomly draw two spikes $6$ hops away from one another and diffuse them as before. We then add normally distributed noise to $b$, varying the levels of Signal-to-Noise-Ratio (SNR). Figure \ref{fig:noise} shows the average and standard deviations, over 32 trials, of the hop error of the solution for different pairs $(\text{SNR}, \theta)$. The solver is robust to noise, given that the error grows smoothly with the noise level. We also confirm that a low diffusion time $\theta$ is important for accurately localizing the sources of the diffusion. This intuitively makes sense, as one can imagine it is hard to infer the past from observations in the distant future.

\vspace{-1.0em}
\subsection{John Snow's Cholera Data}
\vspace{-0.5em}
In 1854, a major outbreak of cholera in the district of Soho, in London, inspired a ground-breaking study by the father of modern epidemiology, physician John Snow \cite{Snow:1855}. Being skeptical of the then dominant miasma theory of diseases, he hypothesized that cholera was transmitted by water, which was reinforced by observing that the infected people's residences seemed to cluster around a water pump on Broad Street.

\begin{figure}[htb]
	\vspace{-1.5em}
	\centering
	\subfloat[Non-zero signal identifying the infected pump (ground-truth).]{\includegraphics[width=0.49\linewidth]{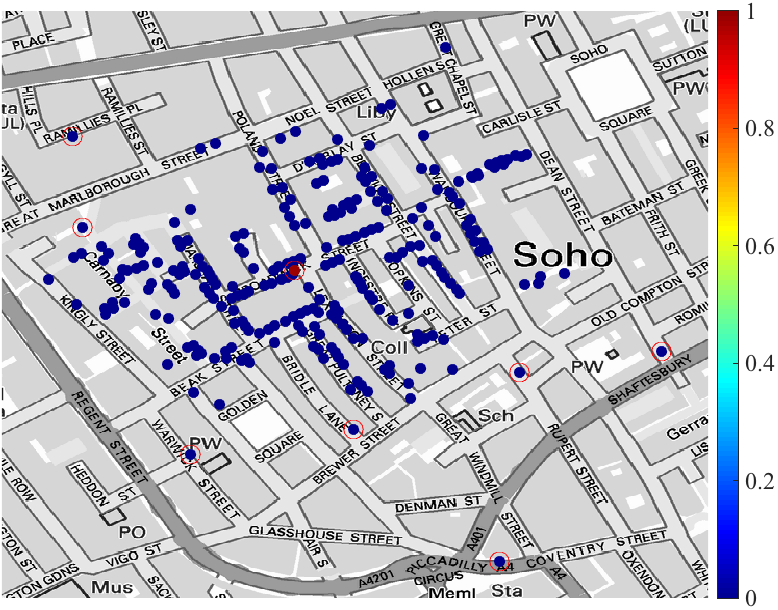}\label{fig:infected_pump}}
    ~
    \subfloat[Observed death count on each node.]{\includegraphics[width=0.49\linewidth]{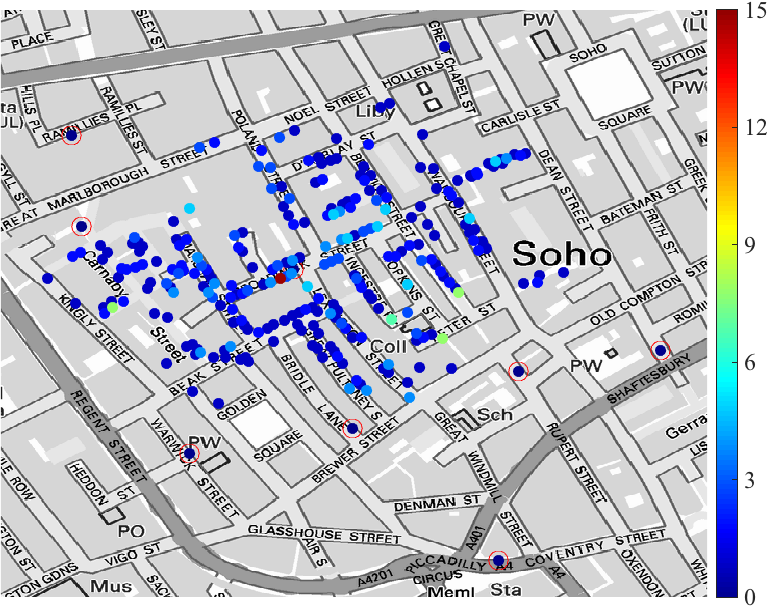}\label{fig:obs_death_count}}
    \vspace{-0.5em}
\caption{John Snow's GIS cholera death data. The red circles indicate the water pumps.}
\label{fig:snow_gis_graph}
\end{figure}

Snow used a map of the region to illustrate how the cases of cholera were distributed around the infected pump, data which was recently converted into modern Geographic Information System (GIS) format and made available in a blog post by Robin Wilson \cite{Wilson:2012}. In this map, there are 250 points marked with deaths by cholera. Each of these points has an associated death count, ranging from 1 to 15, resulting in a total of 489 deaths. There are also 8 special points in the map, corresponding to the closest water pumps in the area during that time. In Figure \ref{fig:infected_pump}, we can see where the infected pump is located, and in Figure \ref{fig:obs_death_count}, we plot the death counts as a heat map. If we make no distinction between water pump nodes and death nodes, can we automatically recover the position of the infected pump? 

We first construct a $k$-NN graph from the data points. The distance between two points is computed as the length of the shortest path between them on the roads of the map. This seems more in tune with the context of the problem than simply computing the Euclidean length of the line segment between the points. The edge weights are given by an exponential kernel similar to the one used for the sensor graph. We assume the number of deaths at each node is a signal that diffuses on the graph according to a linear model. Our goal then is to recover the sparse signal $x$ (source of infection) that generated the observation $b$ (death counts) after being diffused by some $A_{\theta}$ computed from \eqref{eq:heat_kernel}.

\begin{figure}[htb]
	\vspace{-1em}
	\centering
	\includegraphics[width=\linewidth]{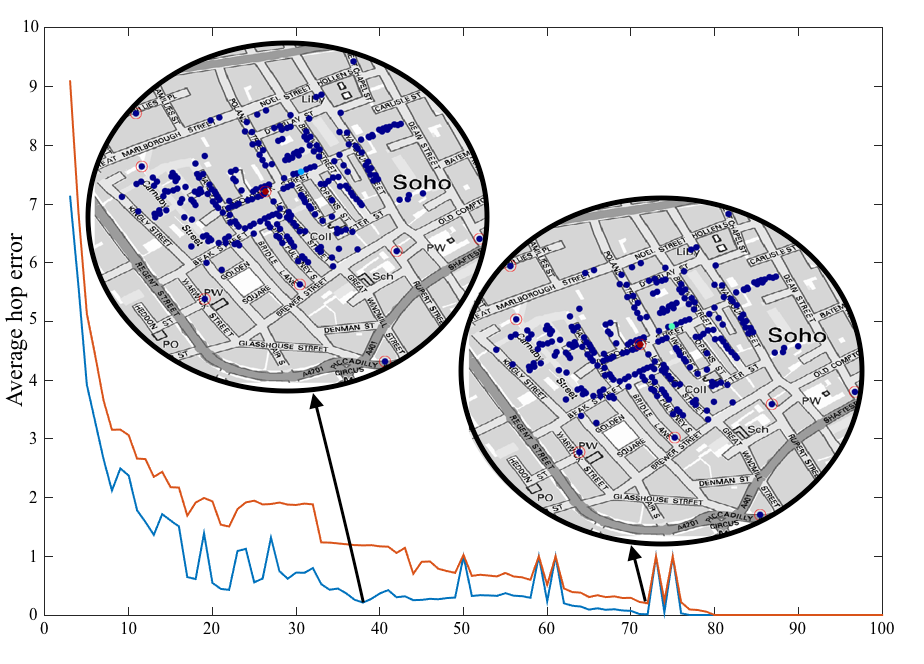}
	\vspace{-1.5em}
    \caption{Snow GIS Graph. Average hop error as a function of the number of neighbors of each node.}
\label{fig:hop_err_neighbors_snow_gis}
\end{figure}

Modeling cholera transmission merely as a heat diffusion is a heavy simplification, which makes it hard to obtain a satisfying source localization when solving the full non-convex problem \eqref{eq:opt_problem}. We have discovered, however, that if we choose a good value for $\theta$ and fix it while solving \eqref{eq:alt_opt_problem}, we manage to reliably recover the location of the infected pump. Another aspect that seemed to noticeably influence the accuracy of the solution was the choice of $k$ when building the $k$-NN graph. In order to investigate that, we ran our solver for varying values of $k$ and observed what was the average hop error of the converging solution in each case. These results can be seen on the blue curve on Figure \ref{fig:hop_err_neighbors_snow_gis}.

\vspace{-0.5em}
 
Even though we manage to satisfyingly identify the infected pump in the above tests (assuming a proper $k$ was chosen), one could argue that simply choosing the node with the maximum death count in Figure \ref{fig:obs_death_count} as the source of infection would guarantee a solution with a small hop error. We can account for this particularity of the data by simply removing this ``outlier'' from our observations $b$. We do this by either multiplying the term inside the $\ell_2$ norm in \eqref{eq:opt_problem} by an observation mask which is zero at the index of the node signal to be removed, or by substituting this outlier by the interpolation of the signal values on the neighboring nodes. Both those strategies have similar results, so we will not bother comparing them here. The orange curve on Figure \ref{fig:hop_err_neighbors_snow_gis} shows the results of the same experiment relative to the blue curve, but this time using the outlier-removed observation. As we can see, the solution is robust to this removal, although one might notice that we need a denser graph to attain similar accuracy levels as before.

\vspace{-1.0em}
\subsection{European Tracer Experiment (ETEX) Data }
\begin{figure}[htb]
	\vspace{-1em}
	\centering
	\subfloat[Non-zero signal identifying the tracer release site (ground-truth).]{\includegraphics[width=0.49\linewidth]{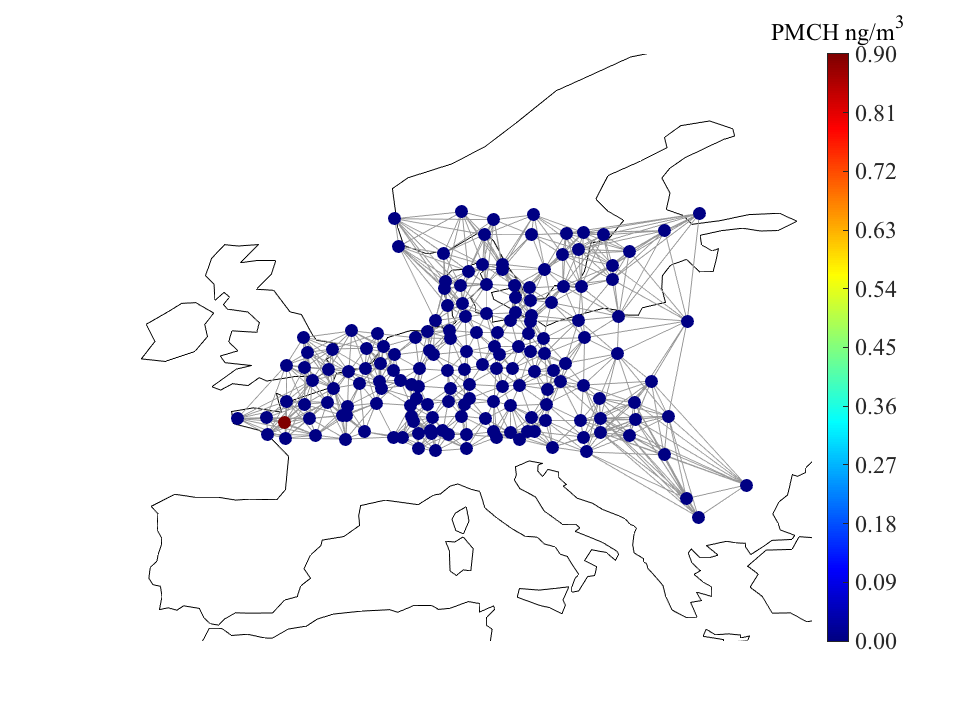}\label{fig:release_site}}
    ~
    \subfloat[Observed cumulative tracer concentration on each node.]{\includegraphics[width=0.49\linewidth]{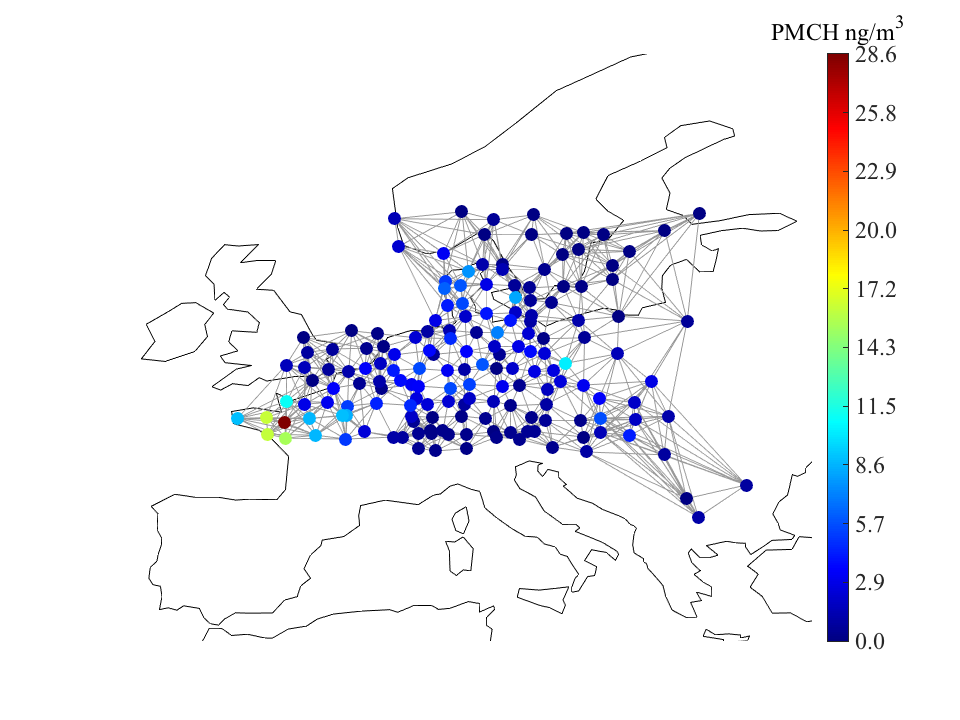}\label{fig:obs_concentration}}
    \vspace{-0.5em}
    \caption{ETEX tracer concentration data.}
\label{fig:etex_graph}
\end{figure}

\vspace{-0.5em}
In the years that followed the Chernobyl accident, the European Community became very interested in monitoring and modeling the atmospheric transport of chemicals, in particular of radionuclides. One of the studies that were devised in this context was the European Tracer Experiment (ETEX) \cite{ETEX:1995}, which took place in 1994. It consisted of two different runs of the same protocol: release, from near Rennes, FR, easily identifiable tracers (perfluorocarbons) on the atmosphere, and sample their concentration, over a period of 72 consecutive hours, at 168 ground-level stations in Western and Eastern Europe. Figure \ref{fig:etex_graph} shows a network of these ground stations, overlaid on a map of Europe. The network is assembled as a k-NN graph, but this time the distance between points is simply computed as the Euclidean distance between their coordinates on the globe.

\begin{figure}[htb]
	\vspace{-1em}
	\centering
	\includegraphics[width=\linewidth]{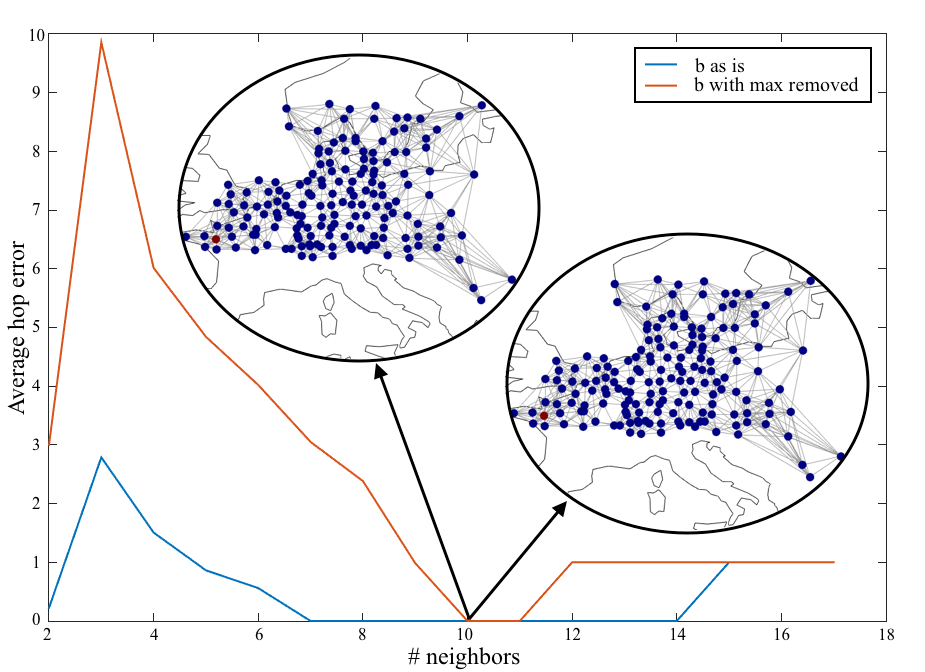}
	\vspace{-1.5em}
    \caption{ETEX Graph. Average hop error as a function of the number of neighbors of each node.}
\label{fig:hop_err_neighbors_etex}
\end{figure}

We model the observations $b$ as the cumulative tracer concentration on each of the nodes. Some stations had invalid samples, or samples that couldn't be quantified. To account for that, we set the concentration values on these nodes as an interpolation of the values on their neighbors. As before, we assume the diffusion is performed by a heat kernel, and fix a given $\theta$ while solving \eqref{eq:opt_problem} only for $x$. The goal is to recover a sparse signal $x$ which is non-zero only at the Rennes station. We were also interested here in observing how the graph construction affects the accuracy of the solution. The blue curve on Figure \ref{fig:hop_err_neighbors_etex} shows the average hop error of the output of our solver for varying graph densities.

The ETEX data also suffers from a similar bias as Snow's data: simply picking the node with the maximum tracer concentration gives us the true source. Once again, we masked or interpolated this value, and ran our solver for different values of $k$ in the graph construction. This is illustrated by the orange curve on Figure \ref{fig:hop_err_neighbors_etex}, and we see that the results are robust to this information removal.

\vspace{-1.0em}
\section{CONCLUSION}
\label{sec:conclusion}
\vspace{-0.5em}
We have introduced a framework for solving source localization problems on graphs that is robust to noise and to source distance, but can be very sensitive to the diffusion time of the heat kernel. This sensitivity can perhaps be attenuated if we allow observations at different time steps throughout the diffusion process. In a future work, perhaps more complex diffusion models could be tested when dealing with real data, in an attempt to get a better behavior from the non-convex problem. For instance, Bertuzzo \textit{et al.} \cite{Bertuzzo:2010fu} develop a fairly detailed dynamic model of cholera epidemics on networks. As a final note, we have also seen that the results depend on the construction of the graph from the given data. Graph construction from data points is still an open problem, but there are some interesting works on the area, e.g., \cite{Friedman:2008df}, \cite{Dong:2015}, \cite{Kalofolias:2016}.

\vspace{-1.0em}
\section{ACKNOWLEDGMENTS}
\vspace{-0.5em}
The first author would like to thank Vassilis Kalofolias for the discussions about the problem and its implementation. The research leading to these results has received funding from the European Union's Seventh Framework Programme (FP7-PEOPLE-2013-ITN) under grant agreement n\textordmasculine ~607290 SpaRTaN.


\vspace{-1.0em}
\bibliographystyle{IEEEbib}
\bibliography{refs}

\begin{thebibliography}{10}

\bibitem{FernandezGranda:2015uc}
C.~Fernandez-Granda,
\newblock ``Super-resolution of point sources via convex programming,''
\newblock in {\em Computational Advances in Multi-Sensor Adaptive Processing
  (CAMSAP), 2015 IEEE 6th International Workshop on}, Dec 2015, pp. 41--44.

\bibitem{Candes:2012uf}
Emmanuel~J Cand{\`e}s and Carlos Fernandez-Granda,
\newblock ``{Towards a Mathematical Theory of Super-resolution},''
\newblock {\em Communications on Pure and Applied Mathematics}, vol. 67, no. 6,
  pp. 906--956, June 2014.

\bibitem{Candes:2013wu}
Emmanuel~J Cand{\`e}s and Carlos Fernandez-Granda,
\newblock ``{Super-Resolution from Noisy Data},''
\newblock {\em Journal of Fourier Analysis and Applications}, vol. 19, no. 6,
  pp. 1229--1254, 2013.

\bibitem{Duval:2015}
Vincent Duval and Gabriel Peyr\'e,
\newblock ``Exact support recovery for sparse spikes deconvolution,''
\newblock {\em Found. Comput. Math.}, vol. 15, no. 5, pp. 1315--1355, Oct.
  2015.

\bibitem{Shuman:2013et}
David~I Shuman, S~K Narang, P~Frossard, A~Ortega, and P~Vandergheynst,
\newblock ``{The emerging field of signal processing on graphs: Extending
  high-dimensional data analysis to networks and other irregular domains},''
\newblock {\em IEEE Signal Processing Magazine}, vol. 30, no. 3, pp. 83--98,
  Apr. 2013.

\bibitem{Pinto:2012tp}
Pedro~C Pinto, Patrick Thiran, and Martin Vetterli,
\newblock ``{Locating the Source of Diffusion in Large-Scale Networks},''
\newblock {\em Physical Review Letters}, vol. 109, no. 6, 2012.

\bibitem{Feizi:2014tb}
Soheil Feizi, Ken Duffy, Manolis Kellis, and Muriel Medard,
\newblock ``Network infusion to infer information sources in networks,''
\newblock Tech. {R}ep. MIT-CSAIL-TR-2014-028, MIT, Dec. 2014.

\bibitem{Zhang:2016wr}
Peng Zhang, Jing He, Guodong Long, Guangyan Huang, and Chengqi Zhang,
\newblock ``{Towards Anomalous Diffusion Sources Detection in a Large
  Network},''
\newblock {\em ACM Transactions on Internet Technology}, vol. 16, Jan. 2016.

\bibitem{Prakash:2012uk}
B~Aditya Prakash, Jilles Vreeken, and Christos Faloutsos,
\newblock ``{Spotting Culprits in Epidemics: How Many and Which Ones?},''
\newblock {\em ICDM}, pp. 11--20, 2012.

\bibitem{Chung:2001ud}
Fan Chung,
\newblock ``{Spectral Graph Theory},''
\newblock vol. 92, Dec. 1996.

\bibitem{Beck:2009:FIS:1658360.1658364}
Amir Beck and Marc Teboulle,
\newblock ``{A Fast Iterative Shrinkage-Thresholding Algorithm for Linear
  Inverse Problems},''
\newblock {\em SIAM J. Img. Sci.}, vol. 2, no. 1, pp. 183--202, Mar. 2009.

\bibitem{Snow:1855}
John Snow,
\newblock ``{Dr. Snow Report},''
\newblock {\em Report on the Cholera Outbreak in the Parish of St. James,
  Westminster, during the Autumn of 1854}, pp. 97--120, July 1855.

\bibitem{Wilson:2012}
Robin Wilson,
\newblock ``{John Snow's famous cholera analysis data in modern GIS formats},''
  2012,
\newblock {[Online]. Available:
  http://blog.rtwilson.com/john-snows-famous-cholera-analysis-data-in-modern-gis-formats/}.

\bibitem{ETEX:1995}
``{European Tracer Experiment (ETEX)},'' 1995,
\newblock {[Online]. Available: https://rem.jrc.ec.europa.eu/RemWeb/etex/}.

\bibitem{Bertuzzo:2010fu}
E~Bertuzzo, R~Casagrandi, M~Gatto, I~Rodriguez-Iturbe, and A~Rinaldo,
\newblock ``{On spatially explicit models of cholera epidemics},''
\newblock {\em Journal of The Royal Society Interface}, vol. 7, no. 43, pp.
  321--333, Feb. 2010.

\bibitem{Friedman:2008df}
Jerome Friedman, Trevor Hastie, and Robert Tibshirani,
\newblock ``{Sparse inverse covariance estimation with the graphical lasso},''
\newblock {\em Biostatistics}, vol. 9, no. 3, pp. 432--441, July 2008.

\bibitem{Dong:2015}
Xiaowen Dong, D.~Thanou, P.~Frossard, and P.~Vandergheynst,
\newblock ``Laplacian matrix learning for smooth graph signal representation,''
\newblock in {\em Acoustics, Speech and Signal Processing (ICASSP), 2015 IEEE
  International Conference on}, April 2015, pp. 3736--3740.

\bibitem{Kalofolias:2016}
Vassilis Kalofolias,
\newblock ``{How to learn a graph from smooth signals},''
\newblock in {\em th International Conference on Artificial Intelligence and
  Statistics AISTATS}, Cadiz, Spain, 2016.

\end{thebibliography}

\end{document}